\documentclass[10pt,twocolumn,letterpaper]{article}

\usepackage[pagenumbers]{cvpr} 

\usepackage[pagebackref,breaklinks,colorlinks,allcolors=cvprblue]{hyperref}
\usepackage{times}
\usepackage{epsfig}
\usepackage{graphicx}
\usepackage{amsmath}
\usepackage{amssymb}
\usepackage{amsmath}
\usepackage{multirow}
\usepackage{booktabs}
\usepackage{algorithm,algpseudocode}
\usepackage{colortbl}
\usepackage{amsthm}
\usepackage{makecell}
\usepackage{ifsym}
\usepackage{marvosym}

\definecolor{mygray}{gray}{.9}

\definecolor{cvprblue}{rgb}{0.21,0.49,0.74}
\definecolor{mypink}{HTML}{FEB5AF}

\newtheorem{definition}{Definition}

\def\paperID{7365} 
\def\confName{ICCV}
\def\confYear{2025}

\title{Efficient Adversarial Training via Criticality-Aware Fine-Tuning}

\author{
    Wenyun Li$^{1,2}$ \quad Zheng Zhang$^{1,2}$\thanks{Corresponding author.}  \quad Dongmei Jiang$^{2}$ \quad Yaowei Wang$^{1,2}$ \quad Xiangyuan Lan$^{2,3}$ \\
    $^{1}$Harbin Institute of Technology, Shenzhen \\
    $^{2}$Pengcheng Laboratory \\
    $^{3}$Pazhou Laboratory (Huangpu) \\
    $\{$liwy, zhangzh, jiangdm, wangyw, lanxy$\}$@pcl.ac.cn
}

\begin{document}
\maketitle
\begin{abstract}
Vision Transformer (ViT) models have achieved remarkable performance across various vision tasks, with scalability being a key advantage when applied to large datasets. This scalability enables ViT models to exhibit strong generalization capabilities. However, as the number of parameters increases, the robustness of ViT models to adversarial examples does not scale proportionally. Adversarial training (AT), one of the most effective methods for enhancing robustness, typically requires fine-tuning the entire model, leading to prohibitively high computational costs, especially for large ViT architectures. In this paper, we aim to robustly fine-tune only a small subset of parameters to achieve robustness comparable to standard AT. To accomplish this, we introduce Criticality-Aware Adversarial Training (CAAT), a novel method that adaptively allocates resources to the most robustness-critical parameters, fine-tuning only selected modules. Specifically, CAAT efficiently identifies parameters that contribute most to adversarial robustness. It then leverages parameter-efficient fine-tuning (PEFT) to robustly adjust weight matrices where the number of critical parameters exceeds a predefined threshold. CAAT exhibits favorable generalization when scaled to larger vision transformer architectures, potentially paving the way for adversarial training at scale, \textit{e.g}, compared with plain adversarial training, CAAT incurs only a 4.3\% decrease in adversarial robustness while tuning approximately 6\% of its parameters. Extensive experiments on three widely used adversarial learning datasets demonstrate that CAAT outperforms state-of-the-art lightweight AT methods with fewer trainable parameters.
\end{abstract}    
\section{Introduction}
\label{sec:intro}

In recent years, the ViT has become a mainstream architecture for various visual tasks, including image classification \cite{dosovitskiy2021an}, image retrieval \cite{10572365}, and object detection \cite{CarionMSUKZ20}. A key advantage behind ViT's success is its scalability when applied to large datasets. The emergent phenomenon is widely regarded as a foundational property of large language models, with similar patterns observed in computer vision \cite{bai2024sequential}.

Despite the significant success of large vision models, their robustness is often overlooked. Even the most advanced ViT models \cite{TouvronCSSJ21,ChenFP21} remain vulnerable to evasion attacks from adversarial examples, where carefully crafted perturbations can easily mislead these models. This poses a serious problem in various applications. Enhancing the robustness of ViT models against adversarial examples to ensure reliable performance is therefore a prominent issue within the security community.
\begin{figure}
    \centering
    \includegraphics[width=0.92\linewidth]{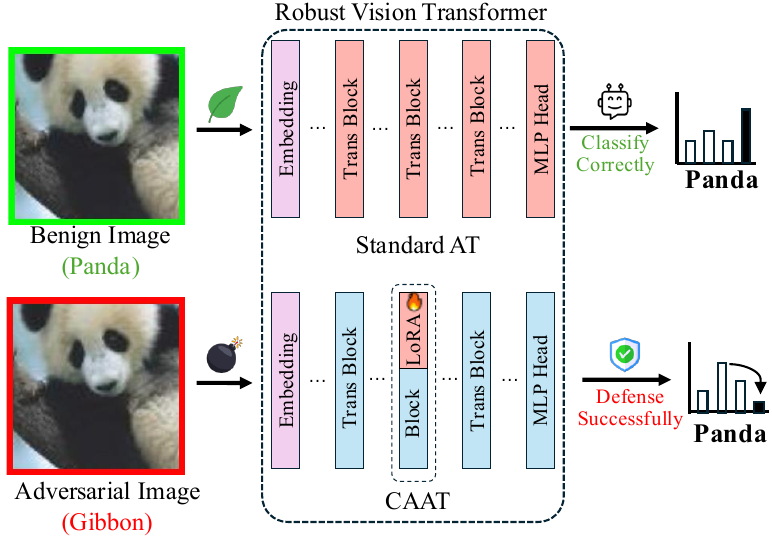}
    \vspace{-.7em}
    \caption{Illustration of standard AT and our CAAT. The parameters highlighted in {\color{cyan}blue} denote frozen parameters, while those in {\color{mypink}rhodo} represent trainable parameters. We only tune the top-$\tau$ critical parameters.}
    \label{fig:nn}
\end{figure}
\begin{figure*}[h]
\centering
\includegraphics[width=1.82\columnwidth]{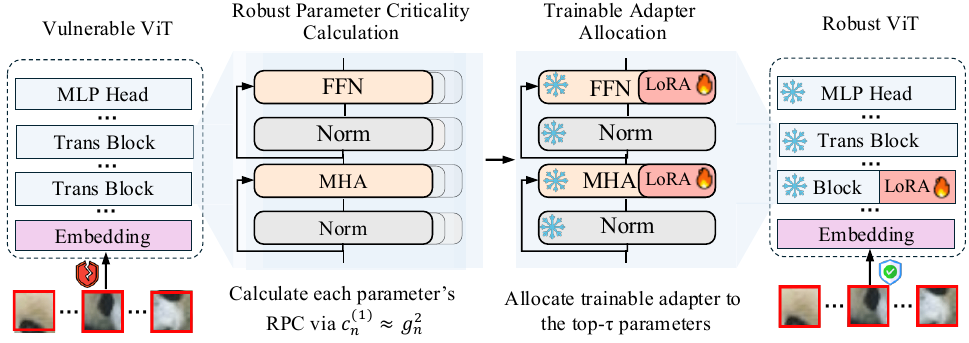}
\vspace{-.7em}
\caption{The pipeline of our proposed Criticality-Aware Adversarial Training. Given a vulnerable ViT, we first identify the top-$\tau$ critical parameters for adversarial robustness. Instead of directly applying AT to these parameters, we leverage the representational  capability of PEFT to enhance adversarial robustness.\label{pipeline}}
\end{figure*}

To improve the robustness of ViT against adversarial examples, researchers have proposed multiple defense methods across preprocessing, training, and postprocessing stages. In the preprocessing phase, adversarial examples in the input data are purified or reduced before being sent to the ViT \cite{NieGHXVA22, Xu0Q18, SongKNEK18}. During the training stage, defenses often involve fine-tuning vulnerable pre-trained ViT models to improve robustness through learning. Notable training-stage defenses include methods from \cite{MadryMSTV18, WuX020}. Finally, postprocessing-stage defenses focus on modifying or analyzing the model’s output \cite{8844598}. Among these methods, adversarial training \cite{MadryMSTV18} is one of the most well-known and effective approaches. Originally introduced by \cite{MadryMSTV18}, AT enhances model robustness by incorporating adversarial examples into the training dataset. Standard AT typically involves fully fine-tuning a pre-trained model, requiring updates to the entire set of model parameters to achieve robustness. The distinction between standard AT and our CAAT is illustrated in Figure.\ref{fig:nn}.

However, for large ViT models, the vast number of parameters makes it infeasible to directly apply traditional AT due to its substantial computational cost. To address the fine-tuning challenges in large models for downstream tasks, recent research has introduced various lightweight parameter-efficient fine-tuning (PEFT) \cite{LesterAC21} methods. Broadly, PEFT methods can be categorized into addition-based and reparameterization-based approaches. Addition-based PEFT methods add new trainable parameters while keeping the original model parameters largely frozen; an example of this is the Adapter\cite{DBLP:conf/icml/HoulsbyGJMLGAG19} method. In contrast, reparameterization-based PEFT methods replace existing model parameters with other trainable parameters rather than adding entirely new ones. Examples of reparameterization-based PEFT include prompt tuning \cite{BrownMRSKDNSSAA20} and LoRA\cite{hu2022lora}. PEFT methods typically require around 1\% of the trainable parameters, improving the performance of pretrained large models on downstream tasks by fine-tuning a small subset of parameters. Recently, FullLoRA-AT\cite{yuan2024fulllora} connected parameter-efficient techniques with AT, using an LNLoRA module to adversarially train a ViT with a lightweight approach. However, FullLoRA-AT\cite{yuan2024fulllora} still tends to tune the backbone globally, leading to challenges with larger trainable parameters and lower robustness performance.

To address the above issues, we introduce a new concept called Robust Parameter Criticality (RPC) to assess the importance of each position in a ViT model for adversarial robustness. RPC quantifies the criticality of each parameter in terms of its contribution to adversarial robustness. We refer to modules with the highest RPC values as robust-critical modules. Our approach allocates trainable parameters to these modules by proposing a novel \textbf{C}riticality-\textbf{A}ware \textbf{A}dversarial \textbf{T}raining (CAAT) scheme, which identifies the most critical positions to adaptively allocate trainable parameters. CAAT consists of two steps: (1) Robust parameter criticality: This step computes the RPC value for each parameter and identifies the robust-critical modules. (2) Trainable Adapter Allocation: Instead of direct tuning, we utilize PEFT techniques to enhance adversarial robustness, training only around 1\% of the parameters. The pipeline of our CAAT is illustrated in Figure.\ref{pipeline}.

Experimental results demonstrate that CAAT achieves comparable robustness performance with approximately 1\% of trainable parameters compared to full fine-tuning in standard AT (see Section \ref{sec:Main_Results}). The computational cost of CAAT has been significantly reduced in terms of both memory usage and training time (Section \ref{sec:Comp_cost}). Furthermore, CAAT demonstrates favorable generalization to larger ViT models (Section \ref{sec:gen}). Additionally, by calculating the RPC values, CAAT provides empirical insights into which parameters in ViT are most susceptible to adversarial examples (see Section \ref{sec:obs_cri}). We further combine CAAT with other AT methods, such as TRADES \cite{zhang2019theoretically}, MART \cite{wang2019improving}, and PRM \cite{mo2022adversarial}, and observe that these incorporations lead to additional robustness improvements (see Section \ref{sec:CorpAT}).

The primary contributions of our research are summarized as follows:
\begin{enumerate}
    \item We propose CAAT to alleviate the computational overhead of standard AT by utilizing PEFT techniques. Specifically, CAAT can achieve nearly the same robustness performance as full adversarial tuning with only 1\% of trainable parameters;
    \item We introduce a novel concept, RPC, to quantify the importance of each parameter for adversarial robustness. In this work, we tune only the most critical parameters to economize computational cost. Additionally, the RPC value serves as a tool for interpreting adversarial robustness;
    \item Extensive experiments validate the superior effectiveness and efficiency of our method compared to other state-of-the-art lightweight AT techniques on three widely used datasets.
\end{enumerate}

\section{Related Work}
\label{sec:formatting}

\subsection{Vision Transformer}
ViT, introduced by \cite{dosovitskiy2021an}, represents a revolutionary architecture following conventional Convolutional Neural Networks (CNNs) \cite{LeCunBBH98, KrizhevskySH12}, which achieved remarkable performance across various visual tasks \cite{kirillov2023segment, cheng2022masked}. ViT captures localized and hierarchical image information through self-attention mechanisms \cite{DBLP:conf/nips/VaswaniSPUJGKP17}. A key advantage of transformer models is their scalability as training data increases, an emergent ability foundational to large language models \cite{dubey2024llama}. Similar phenomena are also observed in computer vision \cite{bai2024sequential}. However, while the discrimination and generalization capabilities of transformer-based models improve, their robustness to abnormal data, such as adversarial examples, remains critical. The most notable robustness enhancement technique, AT \cite{MadryMSTV18}, typically requires fully fine-tuning the target model. The high computational cost has become increasingly prohibitive, particularly given the growing parameter complexity of modern large models. This work pioneers a method to enhance the robustness of transformer-based vision models with a limited number of trainable parameters.
\subsection{Adversarial Training}
DNNs have been widely adopted in various vision tasks \cite{li2023elf, schroff2015facenet} due to their high efficiency and generalization capabilities. However, they still face significant robustness challenges, particularly against adversarial examples \cite{wei2022adversarial}. To enhance robustness, researchers have proposed numerous methods for defense, including adversarial purification \cite{NieGHXVA22}, adversarial training \cite{MadryMSTV18}, and certifiable defenses \cite{croce2019provable}. Among these, AT is widely recognized as the most effective method. First introduced by \cite{MadryMSTV18}, AT improves model robustness by incorporating adversarial examples into the training set. 
TRADES \cite{zhang2019theoretically}, which uses a trade-off loss to balance accuracy and robustness. MART \cite{wang2019improving}, which applies a manifold regularization in the latent space. For adversarial training in ViT models, PRM \cite{mo2022adversarial} enhances robustness by randomly removing gradients in attention blocks during training, while ReiT \cite{gong2024random} employs random entangled self-attention to strengthen adversarial robustness. AAS-AT \cite{jain2024towards} adopts an adaptive attention scaling strategy to improve ViT robustness.

\subsection{Parameter-efficient Fine-tune}
Full fine-tuning is the mainstream approach for adapting large-scale pre-trained models to downstream tasks, wherein all model weights are kept trainable. However, as models grow in size, PEFT \cite{LesterAC21} becomes a more viable option, tuning only a small subset of parameters to meet computational constraints. PEFT methods can be broadly divided into addition-based and reparameterization-based categories. Addition-based PEFT methods add trainable parameters to the model and only adjust these additional components. A well-known addition-based PEFT method is the Adapter\cite{DBLP:conf/icml/HoulsbyGJMLGAG19}, with further developments \cite{ChenGTWSWL22,GaoGZMFZLQ24} enabling larger models to be fine-tuned with minimal additional parameters. However, the attached modules in addition-based PEFT can incur extra computational costs. To address this, reparameterization-based PEFT approaches have been developed. These methods replace original parameters in the model backbone, allowing for efficient fine-tuning without extra modules during inference. The prominent LoRA method \cite{hu2022lora} uses two low-rank matrices that can be merged into the weight matrices. FullLoRA-AT \cite{yuan2024fulllora} was the first to apply PEFT in AT. However, it still tends to adjust the backbone globally. We emphasize the importance of tuning parameters at robustness-critical positions. Identifying such critical parameters through forward propagation offers valuable guidance for more efficient adversarial training. Furthermore, our CAAT approach can determine the most critical backbone positions, providing interpretability regarding how adversarial examples affect DNNs.


\section{Method}

\subsection{Preliminaries}
A $n$-block vision transformer is denoted as $f\left ( \theta  \right ) =LN \left ( x^{\left ( n \right ) }  ; \theta^{\left ( n \right ) }   \right )\circ \dots \circ LN \left ( x^{\left ( 1 \right ) }  ; \theta^{\left ( 1 \right ) }   \right ) $, where $\theta ^{\left ( i \right ) } $ represents the parameters of the $i$-th block, and $LN \left ( \cdot  \right ) $ denotes the layer norm function. We consider the standard classification task with a distribution $\mathcal{D}$ over data points $x \in \mathbb{R} ^d$ and corresponding labels $y \in \left [ k \right ] $. Additionally, a suitable loss function $L (f \left ( x , \theta\right ) ,y)$, for instance, the cross-entropy loss, is always applied to train the ViT. Let $\mathcal{S}$ denote the set of allowed perturbations. For a given data point $x$, an adversarial attack seeks to optimize the perturbation $\delta$ to maximize the empirical risk as follows:
\begin{equation}
    \underset{\delta \in \mathcal{S}  }{max } L\left ( f\left ( x+\delta ,\theta  \right )  ,y   \right ).
\end{equation}
Previous studies \cite{carlini2017towards, GoodfellowSS14} indicate that quasi-indistinguishable adversarial perturbations can easily deceive neural networks. To improve adversarial robustness, AT is commonly used as an effective defense strategy. We define adversarial risk minimization as follows:
\begin{equation}
    \underset{\theta }{min} \rho \left ( \theta  \right ),
\end{equation}
where
\begin{equation}
    \rho \left ( \theta  \right ) =\mathbb{E} _{(x,y)\sim \mathcal{D} }\left [ \underset{\delta  \in \mathcal{S} }{max}L\left ( f\left ( x+\delta ,\theta  \right )  ,y   \right )  \right ] 
\end{equation}
The object of adversarial training is to find model parameters that minimize the adversarial risk. Such notation follows the standard min-max saddle point formulation introduced by \cite{MadryMSTV18}. 
\subsection{Robust Parameter Criticality}
Recent research \cite{zhu2023improving} has shown that the adversarial robustness of fine-tuned backbone parameters varies significantly across different positions. Furthermore, in robust fine-tuning of a pre-trained model, the efficiency of AT can be improved by primarily freezing the pre-trained backbone and only retraining the parameters critical to robustness. Building on this insight, we propose that not all parameters contribute equally to adversarial robustness in AT and introduce a novel criterion to measure the robust criticality of parameters in the pre-trained backbone.
\begin{definition}
(\textbf{Robust Parameter Criticality}) Given the pre-trained model weights $\theta =\left \{ \theta^{\left ( 1 \right ) } ,\theta^{\left ( 2 \right ) }, \dots ,\theta^{\left ( N \right ) } \right \} \in \mathbb{R}^N  $, the robust parameter criticality is denoted as $\mathcal{C}=\left \{ c_1,c_2,\dots ,c_N \right \} $, the criticality $c_n$ for parameter $\theta _n$ is defined as 
\begin{equation}
    c_n=\rho \left ( x+\delta ,\theta  \right ) - \rho \left ( x+\delta ,\theta |\theta_n= \theta_n^{*}\right ) ,
\end{equation}
where $\theta_n^{*}=\underset{\theta_n}{argmin} \left ( \rho \left ( x+\delta ,\theta  \right )  \right ) $. The robust tuned parameters can be reparameterized as $\theta_n^{*}=\theta_n+\Delta _{\theta_n} $, where $\Delta _{\theta_n}$ denotes the update for $\theta_n$ after AT.
\end{definition}

Robust parameter criticality quantifies each parameter’s contribution to the model’s adversarial robustness. Parameters with the lowest criticality are deemed less important, as adjusting their weights has minimal impact on robustness improvement.

The criticality $c_n$ of each parameter can be measured individually. However, this approach becomes intractable as the number of parameters $N$ in a ViT can be very large. For instance, ViT-B \cite{dosovitskiy2021an} contains approximately 85M parameters, making a sequential search for the most critical parameters highly time-consuming.

Inspired by \cite{DBLP:conf/cvpr/MolchanovMTFK19}, we can avoid evaluating all $N$ parameters by approximating $c_n$ in the vicinity of $\theta$ using a first-order Taylor expansion:
\begin{equation}
\label{c_first-order}
c_n^{( 1 )}=-g_n \Delta_{\theta_n},
\end{equation}
where the gradients $\mathbf{g} =\frac{\partial \rho }{\partial \theta }  $ , and $g_n$ representing the $n$-th  elements of the gradient $\mathbf{g}$. The $g_n$ in Eq. \ref{c_first-order} can be easily computed since the gradient $\mathbf{g}$ is already obtained from backpropagation. However, $\Delta _{\theta_n}$ remains unavailable, as it depends on the final convergent value of $\theta_n$. To address this, we are inspired by \cite{DBLP:conf/iclr/CaiZH19},  which suggests using the weight after a single training step as a surrogate for $\theta_n^{*}$. This approximation allows $\Delta _{\theta_n}$ in Eq.\ref{c_first-order} to be computed without requiring full convergence of $\theta_n$. Consequently, Eq.\ref{c_first-order} can be simplified to 
\begin{equation}
    c_n^{( 1 )}=g_n^2 \epsilon ,
\end{equation}
where $\epsilon$ is the learning rate. As $ \epsilon$ is consistent across all parameters, we can omit it when measuring the criticality of each parameter, resulting in the final expression:
\begin{equation}
    c_n^{( 1 )} \approx g_n^2 ,
\end{equation}
Thus, a parameter's criticality can be assessed based on its potential to reduce adversarial risk. The pseudocode for calculating robust parameter criticality is provided in Algorithm.\ref{alg:cap}.

\noindent \textbf{Remark:} The concept is related to previous works such as Taylor Pruning \cite{DBLP:conf/cvpr/MolchanovMTFK19}  and RiFT \cite{zhu2023improving}, though CAAT differs significantly from them in two key aspects. First, the objectives diverge: Taylor Pruning\cite{DBLP:conf/cvpr/MolchanovMTFK19} evaluates parameter importance during training, while RiFT\cite{zhu2023improving} analyzes the impact of a parameter in an adversarially trained model to examine generalization. Second, the implementations vary. Taylor Pruning\cite{DBLP:conf/cvpr/MolchanovMTFK19} estimates importance through the squared change in loss upon removal of a parameter, ultimately deriving importance as $\left ( \mathbf{g}_n \theta _n  \right ) ^2$. RiFT\cite{zhu2023improving}, on the other hand, evaluates a pre-trained adversarial model by adding a perturbation $\Delta _ \theta $ to a parameter $\theta_n$ to analyze the robustness characteristics of module weights.
\begin{algorithm}
\caption{Calculating Robust Parameter Criticality}\label{alg:cap}
\begin{algorithmic}[1]
\Require Pre-trained ViT model with network parameters $\theta$, partial train dataset $\mathcal{D}$.
\Ensure Criticality set $\mathcal{C}=\left \{ c_1,c_2,\dots ,c_N \right \} $.
\State Initialize adversarial dataset: $\mathcal{D}_{adv}=\left \{  \right \} $ and criticality set: $\mathcal{C}=\left \{ 0 \right \}^N $.
\For{Batch $\mathcal{B} \in \mathcal{D}$}  \Comment{Generate adversarial dataset}
\State{$\mathcal{B}^{adv}$ = PGD-10($\theta, \mathcal{B}$)} 
\State{$\mathcal{D}_{adv} = \mathcal{D}_{adv} \bigcup \mathcal{B}^{adv} $}
\EndFor
\For{Batch $\mathcal{B}^{adv} \in \mathcal{D}_{adv}$}  \Comment{Calculate RPC}
\State{Calculate Loss $\rho(\theta, \mathcal{B}^{adv})$} 
\State{Calculate gradients $\textbf{g}$}
\For{$n \in \left \{ 1, \dots, N  \right \} $} 
\State{Update criticality for the $n$-th parameter with: $c_n=c_n+g_n^2$} 
\EndFor
\EndFor
\end{algorithmic}
\end{algorithm}
\subsection{Trainable Adapter Allocation}
Given a robust parameter criticality set 
$\mathcal{C}$ and a parameter critical threshold $\tau$,  an intuitive approach to AT is to directly fine-tune only the top-$\tau$most critical parameters, keeping the remaining parameters fixed—a method we refer to as direct tuning. Specifically, we select the top-$\tau$ critical parameters from $\mathcal{C}$ to form a robust bottleneck parameter set, denoted as $\mathcal{B}$. For a parameter matrix $\Theta  \in \mathbb{R} ^{d_{in} \times d_{out}} $, a binary mask matrix $M \in \mathbb{R} ^{d_{in} \times d_{out}} $ is defined as: 
\begin{equation}
\label{mask}
    M^{i \times j} =\left\{
\begin{aligned}
1, &   &\Theta ^{i \times j}\in \mathcal{B}  \\
0, & &\Theta ^{i \times j} \notin  \mathcal{B}
\end{aligned}
\right.
\end{equation}
where $\Theta ^{i \times j}$ and $M^{i \times j}$ represent the $i \times j$-th elements of $\Theta $ and $M$, respectively. We then train the robust bottleneck parameters using gradient descent, updating the weight matrix according to $\Theta' \gets \Theta - \epsilon \mathbf{g} _{\Theta} \odot M$, where $\mathbf{g} _{\Theta}$ is the gradient of $\Theta$.

However, such direct tuning lacks sufficient representational capability to effectively deal with adversarial example. Given that PEFT \cite{peft} has demonstrated remarkable performance in applications like text-image generation \cite{DBLP:journals/corr/abs-2402-16843} and 
 LLM tuning \cite{ding2023parameter}, achieving these results with less than 1\% of trainable parameters, we propose incorporating PEFT modules into the weight matrices with a high concentration of critical parameters. Specifically, we employ two PEFT techniques in our indirect tuning approach: LoRA \cite{hu2022lora} and Adapter \cite{DBLP:conf/icml/HoulsbyGJMLGAG19}. For example, we apply LoRA \cite{hu2022lora} to the critical weight matrices, where a single update for $\Theta$ can be represented as follows:
\begin{equation}
\Theta ' =\left\{
\begin{aligned}
& \Theta + \Theta_{down}\Theta_{up} ,    &&\mathrm{if} \quad   {\textstyle \sum_{i \times j=0}^{d_{in} \times  d_{out}}} M^{i \times j} \ge \sigma _{tre}    \\
& \Theta - \epsilon \mathbf{g} _{\Theta} \odot M,  &&\mathrm{otherwise}
\end{aligned}
\right.
\end{equation}
where $\Theta_{down} \in \mathbb{R} ^{d_{in}\times d_{r} }  $ and $\Theta_{up} \in \mathbb{R} ^{d_{r}\times d_{out} } $ are two learnable low-rank matrices used to approximate the update of $\Theta$, with $r$ as the rank where $r \ll min\left ( d_{in}, d_{out} \right ) $. Fine-tuning is then applied to $\Theta$ when its number of critical parameters surpasses the threshold hyperparameter $\sigma _{tre}$. For instance, fine-tuning with LoRA requires $2 \times d_{in}\times d_{out}\times d_{r}$ trainable parameters per weight matrix. To ensure that the number of trainable parameters introduced by indirect tuning remains equal to or lower than the number of critical parameters, we set the threshold $\sigma _{tre}$ for LoRA to $2 \times d_{in}\times d_{out}\times d_{r}$.

In this manner, our CAAT framework effectively integrates both direct and indirect tuning methods to enhance the robustness of ViT. As demonstrated in Section.\ref{sec:ablation_study}, empirical results reveal that indirect tuning offers superior robustness performance compared to direct tuning, attributed to its enhanced representational capacity. The full CAAT algorithm is presented in Appendix.

\section{Experiments}
\subsection{Experiment Setup}

\textbf{Datasets} We evaluate our approach on three widely used image classification datasets: CIFAR10 \cite{krizhevsky2009learning}, CIFAR100 \cite{krizhevsky2009learning}, and ImageNet \cite{RussakovskyDSKS15}. CIFAR10 and CIFAR100 contain 60,000 32 × 32 RGB images across 10 and 100 classes, respectively. In our experiments, we split CIFAR10 and CIFAR100 into 50,000 training images and 10,000 test images. ImageNet, corresponding to the ILSVRC 2012 challenge, comprises 1,000 object classes, with 1,281,167 training images and 100,000 testing images.
\begin{table*}[ht]
\small
	\centering
 \caption{Comparisons of different robust fine-tuning methods using a pre-trained ViT-B \cite{dosovitskiy2021an} backbone across various datasets. 'Total params' represents the ratio of the total number of parameters involved in adversarial training, while 'Tuned/Total' denotes the fraction of trainable parameters. The standard adversarial training is highlighted with a gray background. Additionally, the top-1 (\%) accuracy is reported in \textbf{bold}, and the top-2 (\%) accuracy is reported in \underline{underline}. \label{table:table_vit}}
 \vspace{-.7em}
	\begin{tabular}{c | c |c c |c  c c c c  c}
\toprule[0.15em]
 Dataset & Method & \makecell[c]{Tuned params\\(M) } & \makecell[c]{Tuned / Total \\(\%) } & \makecell[c]{Clean Acc \\(\%) } & \makecell[c]{CW-20 \\(\%) } & \makecell[c]{PGD-10 \\(\%) } &  \makecell[c]{AutoAttack \\(\%) } & \makecell[c]{Average \\(\%) } \\
\midrule
\multirow{9}[5]{*}{CIFAR-10} &\cellcolor{mygray} \textsc{Full}  &\cellcolor{mygray} 85.15  & \cellcolor{mygray}100  &\cellcolor{mygray} 82.65    & \cellcolor{mygray}50.69  &\cellcolor{mygray} 53.66  &\cellcolor{mygray} 47.14  &\cellcolor{mygray} 50.50   \\
\cmidrule{2-9}
 & \textsc{Adapter-8} & 1.07 & 1.25 & 80.00  & 48.51 & 47.82 & 44.15 & 46.83  \\
   & \textsc{ Adapter-32} & 1.22     & 1.43      & 80.06    & 48.58  & 47.74     & 42.47 & 46.26  \\
 & \textsc{LoRA-16} & 1.24 & 1.46& 83.24  & 46.71 & 48.96 & 44.96 & 46.88  \\
 & \textsc{Aurora} & 1.15 & 1.35& 81.25  & 45.71 & 43.86 & 42.88 & 44.15  \\
 & \textsc{FullLoRA-AT} & 2.46 & 2.87& 84.93  & 47.67 & 50.30 & 45.28 & 47.75  \\
 & \textsc{HyperAT} & 5.47 & 6.42& 83.70  & 48.73 & \underline{50.46} &  45.32 & 48.17  \\
\cmidrule{2-9}
  & \cellcolor[HTML]{FAEBD7} \textsc{CAAT-Adapter}  & \cellcolor[HTML]{FAEBD7}1.08 & \cellcolor[HTML]{FAEBD7}1.26 & \cellcolor[HTML]{FAEBD7}\underline{85.57} & \cellcolor[HTML]{FAEBD7}\underline{49.11} & \cellcolor[HTML]{FAEBD7}50.15 & \cellcolor[HTML]{FAEBD7}\underline{45.48} & \cellcolor[HTML]{FAEBD7} \makebox[3.6em][r]{\underline{48.25}} \textcolor{red}{\tiny (+1.7)}   \\
  & \cellcolor[HTML]{FAEBD7} \textsc{CAAT-LoRA}  &\cellcolor[HTML]{FAEBD7} 1.09 &\cellcolor[HTML]{FAEBD7} 1.27 & \cellcolor[HTML]{FAEBD7} \textbf{87.12} &\cellcolor[HTML]{FAEBD7} \textbf{49.33} &\cellcolor[HTML]{FAEBD7} \textbf{51.21} &\cellcolor[HTML]{FAEBD7} \textbf{46.37} & \cellcolor[HTML]{FAEBD7} \makebox[3.6em][r]{\textbf{48.97}} \textcolor{red}{\tiny (+1.7)} \\
\midrule
\multirow{9}[5]{*}{CIFAR-100} &\cellcolor{mygray} \textsc{Full}  &\cellcolor{mygray} 85.22  & \cellcolor{mygray}100  &\cellcolor{mygray} 62.01     & \cellcolor{mygray}29.15  &\cellcolor{mygray} 30.56  &\cellcolor{mygray} 27.50  &\cellcolor{mygray} 29.07   \\
\cmidrule{2-9}
 & \textsc{Adapter-8} & 1.07 & 1.26 & 60.01  & 19.57 & 21.35 & 18.63 & 19.85  \\
   & \textsc{ Adapter-32} & 1.22     & 1.43      & 61.17    & 20.88  & 22.79     & 20.36 & 21.34  \\
 & \textsc{LoRA-16} & 1.24 & 1.46&  63.85  & 24.56 & 25.79 & 22.13 & 24.16  \\
 & \textsc{Aurora} & 1.15 & 1.35& 58.67  & 23.44 & 24.03 & 21.55 & 23.01  \\
 & \textsc{FullLoRA-AT} & 2.46 & 2.89& 61.12  & 24.77 & 26.01 & 23.75 & 24.84  \\
 & \textsc{HyperAT} & 5.47 & 6.42& \underline{63.93}   & 24.63 & 25.90 & 23.64 & 24.72  \\
\cmidrule{2-9}
  & \cellcolor[HTML]{FAEBD7} \textsc{CAAT-Adapter}  & \cellcolor[HTML]{FAEBD7}1.07 & \cellcolor[HTML]{FAEBD7}1.26 & \cellcolor[HTML]{FAEBD7}63.15 & \cellcolor[HTML]{FAEBD7}\underline{24.86} & \cellcolor[HTML]{FAEBD7}\underline{26.47} & \cellcolor[HTML]{FAEBD7}\underline{23.87} & \cellcolor[HTML]{FAEBD7}\makebox[3.6em][r]{\underline{25.07}} \textcolor{red}{\tiny (+0.9)}   \\
  & \cellcolor[HTML]{FAEBD7} \textsc{CAAT-LoRA}  &\cellcolor[HTML]{FAEBD7} 1.09 &\cellcolor[HTML]{FAEBD7} 1.28 & \cellcolor[HTML]{FAEBD7}\textbf{64.01} &\cellcolor[HTML]{FAEBD7} \textbf{25.14} &\cellcolor[HTML]{FAEBD7} \textbf{28.05} &\cellcolor[HTML]{FAEBD7} \textbf{24.52} & \cellcolor[HTML]{FAEBD7} \makebox[3.6em][r]{\textbf{25.90}} \textcolor{red}{\tiny (+4.2)}  \\
\midrule
\multirow{9}[5]{*}{ImageNet} &\cellcolor{mygray} \textsc{Full}  &\cellcolor{mygray} 85.15  & \cellcolor{mygray}100  &\cellcolor{mygray} 73.02    & \cellcolor{mygray}55.83  &\cellcolor{mygray} 52.53  &\cellcolor{mygray} 49.72  &\cellcolor{mygray} 52.69   \\
\cmidrule{2-9}
 & \textsc{Adapter-8} & 1.07 & 1.25 & 67.13  & 48.35 & 46.76 & 42.66 & 45.92  \\
   & \textsc{ Adapter-32} & 1.22     & 1.43      & 68.51    & 47.38  & 44.17     & 42.49 & 44.68  \\
 & \textsc{LoRA-16} & 1.24 & 1.46& \underline{72.10}  & 42.78 & 50.90 & 45.95 & 46.54  \\
 & \textsc{Aurora} & 1.18 & 1.39& 70.18  & 50.49 & 50.37 & 45.18 & 48.68  \\
  & \textsc{FullLoRA-AT} & 2.62 & 3.08& 71.23  & 52.33 & 50.95 & 47.10 & 50.13  \\
   & \textsc{HyperAT} & 5.77 & 6.78& 70.33  & \underline{52.88} & 51.33 & 46.88 & 50.36  \\
\cmidrule{2-9}
  & \cellcolor[HTML]{FAEBD7} \textsc{CAAT-Adapter}  & \cellcolor[HTML]{FAEBD7}1.07 & \cellcolor[HTML]{FAEBD7}1.26 & \cellcolor[HTML]{FAEBD7}71.77 & \cellcolor[HTML]{FAEBD7}52.83 & \cellcolor[HTML]{FAEBD7}\underline{51.00} & \cellcolor[HTML]{FAEBD7}\underline{48.01} & \cellcolor[HTML]{FAEBD7} \makebox[3.6em][r]{\underline{50.61}} \textcolor{red}{\tiny (+0.5)}    \\
  & \cellcolor[HTML]{FAEBD7} \textsc{CAAT-LoRA}  &\cellcolor[HTML]{FAEBD7} 1.08 &\cellcolor[HTML]{FAEBD7} 1.27 & \cellcolor[HTML]{FAEBD7}\textbf{72.54} &\cellcolor[HTML]{FAEBD7} \textbf{53.73} &\cellcolor[HTML]{FAEBD7} \textbf{51.56} &\cellcolor[HTML]{FAEBD7} \textbf{49.00} & \cellcolor[HTML]{FAEBD7} \makebox[3.6em][r]{\textbf{51.43}} \textcolor{red}{\tiny (+2.1)}  \\
\bottomrule[0.1em]
\end{tabular}
\end{table*}

\noindent \textbf{Models} We selected the ViT-B \cite{dosovitskiy2021an}, 
ViT-L\cite{dosovitskiy2021an},  ViT-H\cite{dosovitskiy2021an}, Swin-B \cite{liu2021swin} and Swin-L\cite{liu2021swin} models for our experiments.

\noindent \textbf{Evaluation metrics} To evaluate the adversarial robustness of various models, we measure robust test accuracy under different adversarial attack methods. Specifically, we use CW-20 \cite{carlini2017towards}, PGD-10 \cite{MadryMSTV18}, and AutoAttack \cite{DBLP:conf/icml/Croce020a} to assess model robustness. Here, CW-$k$ and PGD-$k$ denote CW and PGD attacks with $k$ iterative steps, respectively. The adversarial budget for these attacks is set to $8/255$, with a step size of $1/255$. Additionally, we evaluate the efficiency of robust fine-tuning for pretrained models by examining the number of learnable parameters required during AT.

\noindent \textbf{Baseline} In our study, we compare CAAT with other PEFT approaches, including Adapter \cite{DBLP:conf/icml/HoulsbyGJMLGAG19}, LoRA \cite{hu2022lora} and Aurora\cite{wang2023paramet}. FullLoRA-AT \cite{yuan2024fulllora} and HyperAT \cite{abs-2410-05951}, which are partial-parameter robust fine-tuning methods aimed at enhancing model robustness. Additionally, we consider the full-parameter  standard AT as a baseline.

\noindent \textbf{Implementation details} For a fair comparison, we set the robust fine-tuning epochs to 30 across all datasets. We employ the AdamW optimizer \cite{loshchilov2017fixing} with cosine learning rate decay. The batch size, learning rate, and weight decay are set to 32, 1$\times$10$^{-3}$, and 1$\times$10$^{-4}$, respectively. Additionally, the number of adversarial samples used to calculate parameter criticality in Algorithm.\ref{alg:cap} is set to 800 for each dataset. All experiments were conducted on a server equipped with 4 NVIDIA A6000 GPUs with 48GB of memory. PyTorch version 2.1.0 was used.

We introduce two variants of our CAAT framework: the addition-based CAAT-Adapter and the reparameterization-based CAAT-LoRA. CAAT-Adapter directly tunes the hidden representations corresponding to sensitive weight matrices, following the method in \cite{DBLP:conf/icml/HoulsbyGJMLGAG19}. In CAAT-LoRA, a product of low-rank matrices is used to approximate updates to the sensitive weight matrices, as described in \cite{hu2022lora}. For both variants, initialization parameters are carefully set according to the original works. Specifically, the bottleneck rank for CAAT-LoRA is set to 16.

\subsection{Main Results}
\label{sec:Main_Results}
In this section, we present the experimental results evaluating the robustness of our CAAT approach against various adversarial attacks on pre-trained models across multiple datasets. To validate the effectiveness of our method, we use CW-20\cite{carlini2017towards}, PGD-10\cite{MadryMSTV18}, and AutoAttack\cite{DBLP:conf/icml/Croce020a} to assess model robustness. The robustness results on the ViT-B and Swin-B models are shown in Table.\ref{table:table_vit} and Appendix. Notably, our proposed CAAT-Adapter and CAAT-LoRA achieve the highest robustness performance with the fewest trainable parameters. For example, CAAT outperforms the state-of-the-art FullLoRA-AT by a clear margin of 1.58\% in mean top-1 accuracy across the three datasets, while using fewer trainable parameters. We attribute our method’s superior robustness to the heuristic selection of critical parameters for fine-tuning. Additionally, we observe that reparameterization-based AT consistently outperforms addition-based variants, indicating that reparameterization-based PEFT is more effective due to its direct modification of the model’s parameters rather than relying on auxiliary parameters.

Furthermore,  our proposed CAAT achieves robustness close to that of full fine-tuning AT, using only approximately 1\% of the parameters. In terms of clean example accuracy, CAAT also demonstrates high performance, indicating that training only a subset of parameters can mitigate catastrophic forgetting during robust fine-tuning. However, when defending against other adversarial attacks, full fine-tuning AT maintains the highest robustness performance, as the frozen parameters in full AT contribute collectively to overall robustness.

\subsection{Computational cost analysis} \label{sec:Comp_cost}
Moreover, we performed an experiment that analyzed the computational cost of CAAT in comparison with full fine-tuning and PEFT methods, such as Adapter and LoRA. Results in Table.\ref{tab:ablation_comp_cost} show that full fine-tuning incurs the highest costs in both memory usage and training time. Addition-based methods like Adapter require more memory during inference. In contrast, our proposed CAAT achieves the lowest resource consumption in both memory and training time.
\begin{table}[!t]
	\centering
 \caption{Cost comparisons with ViT-B backbone on the CIFAR10 robustness when faced PGD-10 attack. The best result is in \textbf{bold}.\label{tab:ablation_comp_cost}}
 \vspace{-.7em}
 \resizebox{\columnwidth}{!}{
 \begin{tabular}{l|ccc}
		\hline
		Method &  \makecell[c]{Inference Memory \\(GB) } & \makecell[c]{Fine-tuning Memory \\(GB) } & \makecell[c]{Train time \\(h) } \\ \hline
  Full &  1.3 & 11.8 &  13.10 \\
  Adapter &  2.0 & 14.0 &  11.30 \\
		LoRA &  1.4 & 10.0 &  10.48 \\ 
  FullLoRA-AT &  1.4 & 8.5 &  8.39 \\
  HyperAT &  1.5 & 9.2 &  8.33 \\
  CAAT &  \textbf{1.3} & \textbf{7.3} &  \textbf{6.00} \\
  \hline
	\end{tabular}
 }
\end{table}
\subsection{Generalization to Larger ViT Architectures}
\label{sec:gen}
Furthermore, our CAAT exhibits favorable generalization when scaled to larger vision transformer architectures, as reported in Table.\ref{tab:arch}. This trend may facilitate the development of adversarially trained, significantly larger vision models, such as ViT-22B \cite{0001DMPHGSCGAJB23}. Notably, we observe that our CAAT relies on only approximately 6\% of trainable parameters, resulting in a mean 4.3\% decrease in adversarial robustness performance.
\begin{table}[!t]
	\centering
 \caption{Adversarial robustness result of our CAAT when scaled to larger vision transformer architecture. The best result is in \textbf{bold}.\label{tab:arch}}
 \vspace{-.7em}
 \resizebox{\columnwidth}{!}{
\begin{tabular}{c|c|c|c|cc}
\hline
Arch                    & Method & Paras & Clean Acc & CW-20 & PGD-10 \\ \hline
\multirow{4}{*}{ViT-L}  &   Full     &   304.33    &     74.82      &  52.77     &    55.39    \\
                        &    LoRA    &   23.58    &    70.53       &    48.90   &    \textbf{54.86}    \\
                        &   FullLoRA-AT     &  81.14     &     61.76      &  50.07     &  44.90      \\
                        &   CAAT     &   \textbf{11.42}    &   \textbf{ 73.72}       &   \textbf{51.49}    &     54.83   \\ \hline
\multirow{4}{*}{ViT-H}  &    Full    &   631.51    &    76.58       &  50.76     &    54.37    \\
                        &   LoRA     &    40.36   &     69.32      &    44.39   &    45.97    \\
                        &   FullLoRA-AT     &    108.39   &   70.33        &   43.60    &   48.33     \\
                        &  CAAT      &    \textbf{27.33}   &     \textbf{71.24}      &    \textbf{46.33}   &     \textbf{50.02}   \\ \hline
\multirow{4}{*}{Swin-L} &    Full    &   196.19    &     75.32      &     47.32  &     50.36   \\
                        &   LoRA     &    18.93   &     44.36      &    33.72   &    35.84    \\
                        &  FullLoRA-AT      &  63.72     &  50.32         &   37.64    &    38.11    \\
                        &   CAAT     &    \textbf{20.66}   &    \textbf{52.37}       &   \textbf{ 40.37}   &    \textbf{41.28}    \\ \hline
\end{tabular}
 }
\end{table}

\subsection{Observations on Criticality Parameters}
\label{sec:obs_cri}
\begin{figure*}
    \begin{subfigure}[b]{0.355\textwidth}
         \centering
         \includegraphics[width=\textwidth]{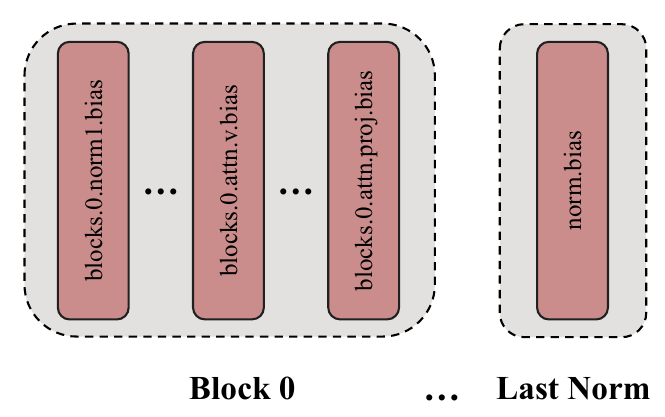}
         \caption{$\tau$=0.05M}
     \end{subfigure}
     \hspace{1.25cm}
     \begin{subfigure}[b]{0.54\textwidth}
         \centering
         \includegraphics[width=\textwidth]{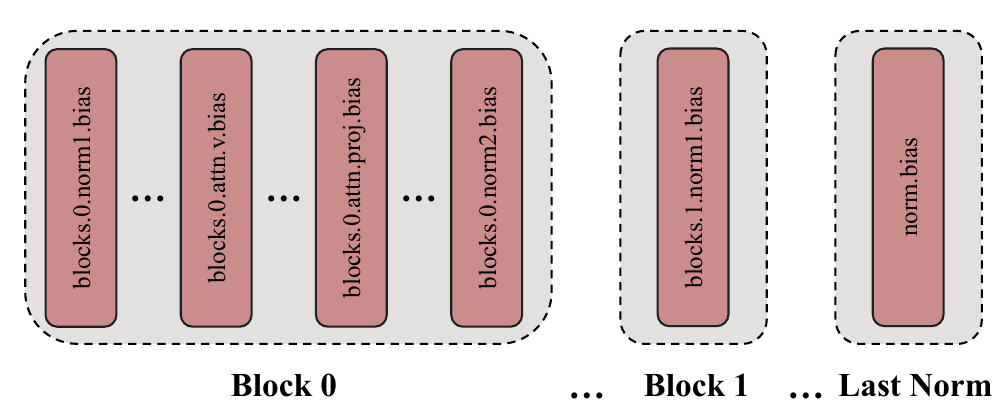}
         \caption{$\tau$=0.1M}
     \end{subfigure} \\
     \begin{subfigure}[b]{\textwidth}
         \centering
         \includegraphics[width=\textwidth]{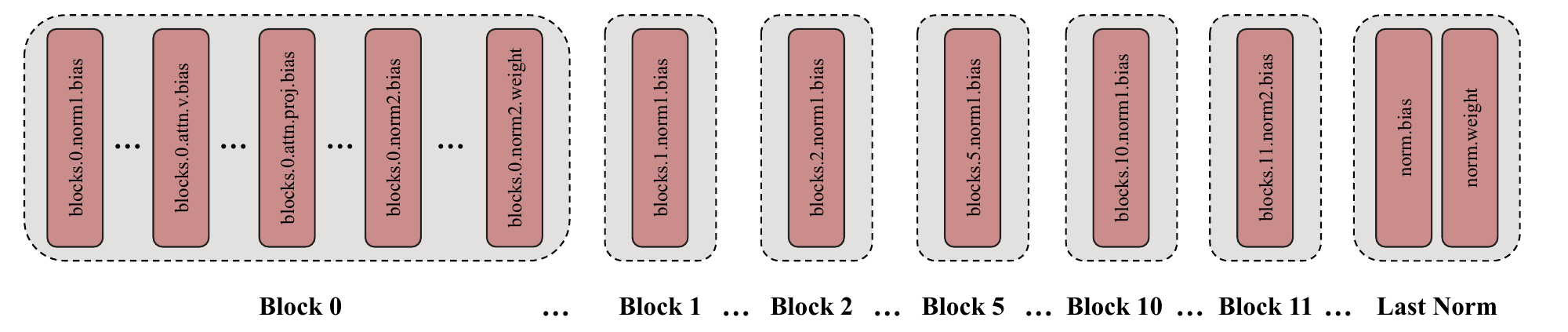}
         \caption{$\tau$=0.2M}
     \end{subfigure}
    \caption{Visualization of the selected critical parameters of ViT-B on CIFAR10 when subjected to a PGD-10 attack under 0.05M, 0.1M, and 0.2M parameter budgets, respectively.}
    \label{fig:critcal_nn}
\end{figure*}

Our criticality criterion identifies the most vulnerable parameters to adversarial examples, serving as a tool to interpret neural network susceptibility to such attacks. In Figure. \ref{fig:critcal_nn},  we visualize the selected critical parameters of the pre-trained ViT-B backbone when exposed to PGD-10 attacks with trainable parameter budgets of 0.05M, 0.1M, and 0.2M, respectively. 

From Figure \ref{fig:critcal_nn}, we observe that under extremely limited parameter budgets, \textit{e.g.}, $\tau$=0.05, only the first block and the final normalization layer are selected, indicating that adversarial examples primarily affect the initial block and the final normalization layer. As the trainable parameter budget increases, the normalization layers in the feed-forward network of the middle blocks gradually enter the critical set, suggesting that, at this stage, fully connected layers gain importance in terms of vulnerability.

\subsection{Corporation with other AT methods}
\label{sec:CorpAT}
To further validate the effectiveness of CAAT, we incorporated it with additional SOTA AT methods to enhance robustness using a minimal number of trainable parameters. We conducted experiments with ViT-B trained on CIFAR10 and CIFAR100 using three different AT methods: TRADES \cite{zhang2019theoretically}, MART \cite{wang2019improving}, and PRM \cite{mo2022adversarial}. The results, reported in Table. \ref{table:cort_with_at}, show that when combined with other state-of-the-art AT methods, CAAT significantly outperforms other partial-parameter fine-tuning methods, such as FullLoRA-AT, in terms of robustness across different datasets.
\begin{table}[]
\centering
 \caption{Results of combining CAAT with other AT methods. The best result is in \textbf{bold}. \label{table:cort_with_at}}
 \vspace{-.7em}
\resizebox{\columnwidth}{!}{\begin{tabular}{c|c|c|c|ccc}
\hline
           Dataset               &         \makecell[c]{Combined\\ AT }           & Method  & \makecell[c]{Trainable\\ Params(M) }  &            \makecell[c]{Clean Acc \\(\%) }           &                      \makecell[c]{PGD-10 \\(\%) }           &    \makecell[c]{AutoAttack \\(\%) }                   \\ \hline
\multirow{12}{*}{CIFAR-10} & \multirow{4}{*}{TRADES} & Full & 85.15 &  83.57                              &          53.59             &           48.99            \\ 
                          &                    & LoRA & 4.72 &       84.73                                  &         50.45              &          45.90             \\
                          &                    & FullLoRA-AT  & 4.86 & 86.88                                          &            51.66           &        46.01               \\
                          &                    & CAAT & 1.10 & \textbf{86.95}                     &              \textbf{52.15}          &         \textbf{47.60}              \\ \cline{2-7} 
                          & \multirow{4}{*}{MART} & Full & 85.15 &  83.66                     &                       53.57             &           49.05            \\  
                          &                    & LoRA & 4.72 &       84.80                                  &         50.69              &          46.04             \\
                          &                    & FullLoRA-AT  & 4.86 & \textbf{87.55}                     &             52.04           &       47.15                \\
                          &                    & CAAT & 1.10 & 87.72                                          &        \textbf{52.21}               &          \textbf{47.40}             \\ \cline{2-7} 
                          & \multirow{4}{*}{PRM} & Full & 85.15 &         83.45              &                      53.57           &     48.76                  \\  
                          &                    & LoRA & 4.72 &       84.61                                   &             50.05          &        45.67               \\
                          &                    & FullLoRA-AT  & 4.86 & 86.71                                       &           51.27            &        47.29               \\
                          &                    & CAAT & 1.10 & \textbf{86.87}                    &            \textbf{52.01}           &            \textbf{47.58}           \\ \hline
\multirow{12}{*}{CIFAR-100} & \multirow{4}{*}{TRADES} & Full & 85.22 &    61.73                                &        31.27               &         28.15              \\ 
                          &                    & LoRA & 4.72 &          61.83                                &              26.33         &             23.09          \\
                          &                    & FullLoRA-AT  & 4.86 & 61.88                                       &          28.93             &       26.44                \\
                          &                    & CAAT & 1.10 &\textbf{ 62.03}                     &              \textbf{29.01 }         &           \textbf{26.55}            \\ \cline{2-7}  
                          & \multirow{4}{*}{MART} & Full & 85.22 &    61.80                   &                        31.32          &  28.26                     \\ 
                          &                    & LoRA & 4.72 &          61.67                                 &           26.50            &         23.40              \\
                          &                    & FullLoRA-AT  & 4.86 & 62.00                                         &           29.20            &     26.59                  \\
                          &                    & CAAT & 1.10 & \textbf{62.05}                     &              \textbf{29.25}          &           \textbf{26.67}            \\ \cline{2-7}  
                          & \multirow{4}{*}{PRM} & Full & 85.22 &    61.40                   &                       31.06            &    28.07                   \\ 
                          &                    & LoRA & 4.72 &           62.53                                &         26.17              &           22.86            \\
                          &                    & FullLoRA-AT  & 4.86 & 61.70                                          &          28.75             &      26.32                 \\
                          &                    & CAAT & 1.10 & \textbf{61.80}                     &            \textbf{28.87}            &           \textbf{26.36 }           \\ \hline
\end{tabular}}
\end{table}

\subsection{Ablation Study}
\label{sec:ablation_study}
\begin{figure}[h]
\centering
\includegraphics[width=0.75\columnwidth]{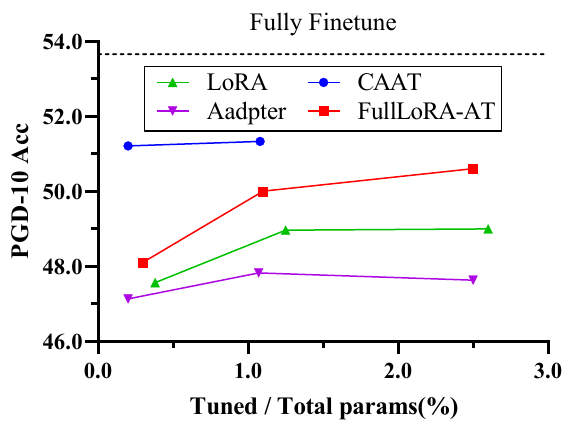}
\vspace{-.7em}
\caption{Accuracy vs. parameter efficiency of ViT-B on the CIFAR10 dataset under PGD-10 attack defense.\label{acc_vs_para}}
\end{figure}

\textbf{Effect of the trainable parameter proportion} We conduct the effectiveness of our proportion of trainable parameters to determine its impact on model robustness. The "Tuned/Total" denotes the fraction of trainable parameters. The result is shown in Figure. \ref{acc_vs_para}. The dashline indicates the fully fine-tuned upper bound. From the figure, we can find that CAAT can outperform the other methods in model robustness while being more parameter efficient.

\noindent \textbf{Effect of direct and indirect tuning} We investigate the effect of direct and indirect tuning on model robustness. Direct tuning involves directly fine-tuning critical parameters in response to adversarial examples, while indirect tuning employs partial parameter adjustments using PEFT techniques such as LoRA and Adapter. We conducted experiments with ViT-B on the CIFAR10 dataset under the PGD-10 attack. The results are presented in Table. \ref{tab:direct}, indicate that indirect tuning provides superior robustness overall.

\begin{table}[]
	\centering
 \caption{Ablation study on direct and indirect tuning. The best result is in \textbf{bold}.\label{tab:direct}}
 \resizebox{\columnwidth}{!}{
 \begin{tabular}{l|ccccc}
		\hline
		Method & \makecell[c]{Tuned / Total \\(\%) } & \makecell[c]{Clean Acc \\(\%) } & \makecell[c]{CW-20 \\(\%) } & \makecell[c]{PGD-10 \\(\%) } & \makecell[c]{AutoAttack \\(\%) } \\ \hline
		LoRA & 1.24 & 83.24 & 46.71 & 50.30 & 45.28 \\
		CAAT(direct tuning) & 1.15 & 79.68 & 46.82 & 50.35 & 45.55 \\
		CAAT(indirect tuning) & 1.09 & \textbf{87.12} & \textbf{49.33} & \textbf{51.21} & \textbf{46.37} \\ \hline
	\end{tabular}
 }
\end{table}

\noindent \textbf{Effect of different rank $r$} We evaluate the robustness of our proposed CAAT-LoRA across different ranks $r$ in the allocated LoRA. As shown in Table. \ref{tab:rank}, the model achieves the highest adversarial accuracy at ranks $r=16$ and $r=32$. Considering the trade-off between parameter efficiency and model robustness, we set $r=16$ as the default configuration in our work.
\begin{table}[h]
\footnotesize
\centering
  \caption{\label{tab:rank}%
     Results of our CAAT under different rank $r$.
  }
  \vspace{-0.2cm}
\begin{tabular}{c|c|c|cccc}
\hline
Rank $r$ & Paras & Clean Acc & CW-20 & PGD-10 & AutoAttack \\ \hline
8      & 0.57            & 82.06     & 46.32 & 49.64  & 43.1       \\
16     & 1.09            & \textbf{87.12}     & 49.33 & \textbf{51.21}  & 46.37      \\
32     & 2.34            & 86.77     & \textbf{50.15} & 50.38  & \textbf{46.94}      \\
64     & 4.93            & 84.25     & 47.66 & 50.5   & 45.86      \\ \hline
\end{tabular}
\end{table}

\section{Conclusion}
In this paper, we propose a lightweight CAAT method to efficiently enhance the adversarial robustness of pretrained ViT models using a partial critical-aware PEFT technique. By identifying the most critical parameters for adversarial robustness, our CAAT significantly reduces the number of trainable parameters in ViT. Furthermore, the robustness of CAAT benefits from its extensive representational ability. Extensive experiments demonstrate that our method not only requires less computational budget compared to previous approaches but also achieves significant improvements in the robustness of models against adversarial attacks. In the future, we plan to extend our method to larger multimodal models, such as LLaVA \cite{abs-2408-03326}, to further enhance the adversarial robustness of large-scale transformer models.
{
    \small
    \bibliographystyle{IEEEtran}
    \bibliography{main}

\begin{thebibliography}{10}
\providecommand{\url}[1]{#1}
\csname url@samestyle\endcsname
\providecommand{\newblock}{\relax}
\providecommand{\bibinfo}[2]{#2}
\providecommand{\BIBentrySTDinterwordspacing}{\spaceskip=0pt\relax}
\providecommand{\BIBentryALTinterwordstretchfactor}{4}
\providecommand{\BIBentryALTinterwordspacing}{\spaceskip=\fontdimen2\font plus
\BIBentryALTinterwordstretchfactor\fontdimen3\font minus \fontdimen4\font\relax}
\providecommand{\BIBforeignlanguage}[2]{{%
\expandafter\ifx\csname l@#1\endcsname\relax
\typeout{** WARNING: IEEEtran.bst: No hyphenation pattern has been}%
\typeout{** loaded for the language `#1'. Using the pattern for}%
\typeout{** the default language instead.}%
\else
\language=\csname l@#1\endcsname
\fi
#2}}
\providecommand{\BIBdecl}{\relax}
\BIBdecl

\bibitem{dosovitskiy2021an}
\BIBentryALTinterwordspacing
A.~Dosovitskiy, L.~Beyer \emph{et~al.}, ``An image is worth 16x16 words: Transformers for image recognition at scale,'' in \emph{International Conference on Learning Representations}, 2021. [Online]. Available: \url{https://openreview.net/forum?id=YicbFdNTTy}
\BIBentrySTDinterwordspacing

\bibitem{10572365}
Q.~Wu, Z.~Zhang, Y.~Liu, J.~Zhang, and L.~Nie, ``Contrastive multi-bit collaborative learning for deep cross-modal hashing,'' \emph{IEEE Transactions on Knowledge and Data Engineering}, vol.~36, no.~11, pp. 5835--5848, 2024.

\bibitem{CarionMSUKZ20}
\BIBentryALTinterwordspacing
N.~Carion, F.~Massa, G.~Synnaeve, N.~Usunier, A.~Kirillov, and S.~Zagoruyko, ``End-to-end object detection with transformers,'' in \emph{ECCV}, ser. Lecture Notes in Computer Science, A.~Vedaldi, H.~Bischof, T.~Brox, and J.~Frahm, Eds., vol. 12346.\hskip 1em plus 0.5em minus 0.4em\relax Springer, 2020, pp. 213--229. [Online]. Available: \url{https://doi.org/10.1007/978-3-030-58452-8\_13}
\BIBentrySTDinterwordspacing

\bibitem{bai2024sequential}
Y.~Bai, X.~Geng, K.~Mangalam, A.~Bar, A.~L. Yuille, T.~Darrell, J.~Malik, and A.~A. Efros, ``Sequential modeling enables scalable learning for large vision models,'' in \emph{Proceedings of the IEEE/CVF Conference on Computer Vision and Pattern Recognition}, 2024, pp. 22\,861--22\,872.

\bibitem{TouvronCSSJ21}
\BIBentryALTinterwordspacing
H.~Touvron, M.~Cord, A.~Sablayrolles, G.~Synnaeve, and H.~J{\'{e}}gou, ``Going deeper with image transformers,'' in \emph{{ICCV}}, 2021, pp. 32--42. [Online]. Available: \url{https://doi.org/10.1109/ICCV48922.2021.00010}
\BIBentrySTDinterwordspacing

\bibitem{ChenFP21}
\BIBentryALTinterwordspacing
C.~R. Chen, Q.~Fan, and R.~Panda, ``Crossvit: Cross-attention multi-scale vision transformer for image classification,'' in \emph{{ICCV}}, 2021, pp. 347--356. [Online]. Available: \url{https://doi.org/10.1109/ICCV48922.2021.00041}
\BIBentrySTDinterwordspacing

\bibitem{NieGHXVA22}
\BIBentryALTinterwordspacing
W.~Nie, B.~Guo, Y.~Huang, C.~Xiao, A.~Vahdat, and A.~Anandkumar, ``Diffusion models for adversarial purification,'' in \emph{International Conference on Machine Learning}, ser. Proceedings of Machine Learning Research, K.~Chaudhuri, S.~Jegelka, L.~Song, C.~Szepesv{\'{a}}ri, G.~Niu, and S.~Sabato, Eds., vol. 162.\hskip 1em plus 0.5em minus 0.4em\relax {PMLR}, 2022, pp. 16\,805--16\,827. [Online]. Available: \url{https://proceedings.mlr.press/v162/nie22a.html}
\BIBentrySTDinterwordspacing

\bibitem{Xu0Q18}
\BIBentryALTinterwordspacing
W.~Xu, D.~Evans, and Y.~Qi, ``Feature squeezing: Detecting adversarial examples in deep neural networks,'' in \emph{Network and Distributed System Security Symposium}, 2018. [Online]. Available: \url{https://www.ndss-symposium.org/wp-content/uploads/2018/02/ndss2018\_03A-4\_Xu\_paper.pdf}
\BIBentrySTDinterwordspacing

\bibitem{SongKNEK18}
\BIBentryALTinterwordspacing
Y.~Song, T.~Kim, S.~Nowozin, S.~Ermon, and N.~Kushman, ``Pixeldefend: Leveraging generative models to understand and defend against adversarial examples,'' in \emph{International Conference on Learning Representations}, 2018. [Online]. Available: \url{https://openreview.net/forum?id=rJUYGxbCW}
\BIBentrySTDinterwordspacing

\bibitem{MadryMSTV18}
A.~Madry, A.~Makelov, L.~Schmidt, D.~Tsipras, and A.~Vladu, ``Towards deep learning models resistant to adversarial attacks,'' in \emph{International Conference on Learning Representations}, 2018.

\bibitem{WuX020}
\BIBentryALTinterwordspacing
D.~Wu, S.~Xia, and Y.~Wang, ``Adversarial weight perturbation helps robust generalization,'' in \emph{Neural Information Processing Systems}, H.~Larochelle, M.~Ranzato, R.~Hadsell, M.~Balcan, and H.~Lin, Eds., 2020. [Online]. Available: \url{https://proceedings.neurips.cc/paper/2020/hash/1ef91c212e30e14bf125e9374262401f-Abstract.html}
\BIBentrySTDinterwordspacing

\bibitem{8844598}
T.~Lee, B.~Edwards, I.~Molloy, and D.~Su, ``Defending against neural network model stealing attacks using deceptive perturbations,'' in \emph{2019 IEEE Security and Privacy Workshops (SPW)}, 2019, pp. 43--49.

\bibitem{LesterAC21}
\BIBentryALTinterwordspacing
B.~Lester, R.~Al{-}Rfou, and N.~Constant, ``The power of scale for parameter-efficient prompt tuning,'' in \emph{Empirical Methods in Natural Language Processing}, M.~Moens, X.~Huang, L.~Specia, and S.~W. Yih, Eds.\hskip 1em plus 0.5em minus 0.4em\relax Association for Computational Linguistics, 2021, pp. 3045--3059. [Online]. Available: \url{https://doi.org/10.18653/v1/2021.emnlp-main.243}
\BIBentrySTDinterwordspacing

\bibitem{DBLP:conf/icml/HoulsbyGJMLGAG19}
\BIBentryALTinterwordspacing
N.~Houlsby, A.~Giurgiu, S.~Jastrzebski, B.~Morrone, Q.~de~Laroussilhe, A.~Gesmundo, M.~Attariyan, and S.~Gelly, ``Parameter-efficient transfer learning for {NLP},'' in \emph{Proceedings of International Conference on Machine Learning}, ser. Proceedings of Machine Learning Research, K.~Chaudhuri and R.~Salakhutdinov, Eds., vol.~97, 2019, pp. 2790--2799. [Online]. Available: \url{http://proceedings.mlr.press/v97/houlsby19a.html}
\BIBentrySTDinterwordspacing

\bibitem{BrownMRSKDNSSAA20}
\BIBentryALTinterwordspacing
T.~B. Brown, B.~Mann, N.~Ryder, M.~Subbiah \emph{et~al.}, ``Language models are few-shot learners,'' in \emph{Neural Information Processing Systems}, H.~Larochelle, M.~Ranzato, R.~Hadsell, M.~Balcan, and H.~Lin, Eds., 2020. [Online]. Available: \url{https://proceedings.neurips.cc/paper/2020/hash/1457c0d6bfcb4967418bfb8ac142f64a-Abstract.html}
\BIBentrySTDinterwordspacing

\bibitem{hu2022lora}
E.~J. Hu, Y.~Shen, P.~Wallis, Z.~Allen-Zhu, Y.~Li, S.~Wang, L.~Wang, and W.~Chen, ``Lo{RA}: Low-rank adaptation of large language models,'' in \emph{International Conference on Learning Representations}, 2022.

\bibitem{yuan2024fulllora}
Z.~Yuan, J.~Zhang, and S.~Shan, ``Fulllora-at: Efficiently boosting the robustness of pretrained vision transformers,'' \emph{arXiv preprint arXiv:2401.01752}, 2024.

\bibitem{zhang2019theoretically}
H.~Zhang, Y.~Yu, J.~Jiao, E.~Xing, L.~El~Ghaoui, and M.~Jordan, ``Theoretically principled trade-off between robustness and accuracy,'' in \emph{International conference on machine learning}.\hskip 1em plus 0.5em minus 0.4em\relax PMLR, 2019, pp. 7472--7482.

\bibitem{wang2019improving}
Y.~Wang, D.~Zou, J.~Yi, J.~Bailey, X.~Ma, and Q.~Gu, ``Improving adversarial robustness requires revisiting misclassified examples,'' in \emph{International conference on learning representations}, 2019.

\bibitem{mo2022adversarial}
Y.~Mo, D.~Wu, Y.~Wang, Y.~Guo, and Y.~Wang, ``When adversarial training meets vision transformers: Recipes from training to architecture,'' \emph{Advances in Neural Information Processing Systems}, vol.~35, pp. 18\,599--18\,611, 2022.

\bibitem{LeCunBBH98}
\BIBentryALTinterwordspacing
Y.~LeCun, L.~Bottou, Y.~Bengio, and P.~Haffner, ``Gradient-based learning applied to document recognition,'' \emph{Proc. {IEEE}}, vol.~86, no.~11, pp. 2278--2324, 1998. [Online]. Available: \url{https://doi.org/10.1109/5.726791}
\BIBentrySTDinterwordspacing

\bibitem{KrizhevskySH12}
\BIBentryALTinterwordspacing
A.~Krizhevsky, I.~Sutskever, and G.~E. Hinton, ``Imagenet classification with deep convolutional neural networks,'' in \emph{Advances in Neural Information Processing Systems}, P.~L. Bartlett, F.~C.~N. Pereira, C.~J.~C. Burges, L.~Bottou, and K.~Q. Weinberger, Eds., 2012, pp. 1106--1114. [Online]. Available: \url{https://proceedings.neurips.cc/paper/2012/hash/c399862d3b9d6b76c8436e924a68c45b-Abstract.html}
\BIBentrySTDinterwordspacing

\bibitem{kirillov2023segment}
A.~Kirillov, E.~Mintun, N.~Ravi \emph{et~al.}, ``Segment anything,'' in \emph{Proceedings of the IEEE/CVF International Conference on Computer Vision}, 2023, pp. 4015--4026.

\bibitem{cheng2022masked}
B.~Cheng, I.~Misra, A.~G. Schwing, A.~Kirillov, and R.~Girdhar, ``Masked-attention mask transformer for universal image segmentation,'' in \emph{Proceedings of the IEEE/CVF conference on computer vision and pattern recognition}, 2022, pp. 1290--1299.

\bibitem{DBLP:conf/nips/VaswaniSPUJGKP17}
\BIBentryALTinterwordspacing
A.~Vaswani, N.~Shazeer, N.~Parmar, J.~Uszkoreit, L.~Jones, A.~N. Gomez, L.~Kaiser, and I.~Polosukhin, ``Attention is all you need,'' in \emph{Advances in Neural Information Processing Systems}, I.~Guyon, U.~von Luxburg, S.~Bengio, H.~M. Wallach, R.~Fergus, S.~V.~N. Vishwanathan, and R.~Garnett, Eds., 2017, pp. 5998--6008. [Online]. Available: \url{https://proceedings.neurips.cc/paper/2017/hash/3f5ee243547dee91fbd053c1c4a845aa-Abstract.html}
\BIBentrySTDinterwordspacing

\bibitem{dubey2024llama}
A.~Dubey, A.~Jauhri, A.~Pandey \emph{et~al.}, ``The llama 3 herd of models,'' \emph{arXiv preprint arXiv:2407.21783}, 2024.

\bibitem{li2023elf}
W.~Li and C.-M. Pun, ``Elf: An end-to-end local and global multimodal fusion framework for glaucoma grading,'' in \emph{2023 IEEE International Conference on Bioinformatics and Biomedicine (BIBM)}.\hskip 1em plus 0.5em minus 0.4em\relax IEEE, 2023, pp. 4081--4085.

\bibitem{schroff2015facenet}
F.~Schroff, D.~Kalenichenko, and J.~Philbin, ``Facenet: A unified embedding for face recognition and clustering,'' in \emph{Proceedings of the IEEE conference on computer vision and pattern recognition}, 2015, pp. 815--823.

\bibitem{wei2022adversarial}
X.~Wei, Y.~Guo, and J.~Yu, ``Adversarial sticker: A stealthy attack method in the physical world,'' \emph{IEEE Transactions on Pattern Analysis and Machine Intelligence}, vol.~45, no.~3, pp. 2711--2725, 2022.

\bibitem{croce2019provable}
F.~Croce and M.~Hein, ``Provable robustness against all adversarial $ l_p $-perturbations for $ p \geq 1$,'' \emph{arXiv preprint arXiv:1905.11213}, 2019.

\bibitem{gong2024random}
H.~Gong, M.~Dong, S.~Ma, S.~Camtepe, S.~Nepal, and C.~Xu, ``Random entangled tokens for adversarially robust vision transformer,'' in \emph{Proceedings of the IEEE/CVF Conference on Computer Vision and Pattern Recognition}, 2024, pp. 24\,554--24\,563.

\bibitem{jain2024towards}
S.~Jain and T.~Dutta, ``Towards understanding and improving adversarial robustness of vision transformers,'' in \emph{Proceedings of the IEEE/CVF Conference on Computer Vision and Pattern Recognition}, 2024, pp. 24\,736--24\,745.

\bibitem{ChenGTWSWL22}
\BIBentryALTinterwordspacing
S.~Chen, C.~Ge, Z.~Tong, J.~Wang, Y.~Song, J.~Wang, and P.~Luo, ``Adaptformer: Adapting vision transformers for scalable visual recognition,'' in \emph{Neural Information Processing Systems}, S.~Koyejo, S.~Mohamed, A.~Agarwal, D.~Belgrave, K.~Cho, and A.~Oh, Eds., 2022. [Online]. Available: \url{http://papers.nips.cc/paper\_files/paper/2022/hash/69e2f49ab0837b71b0e0cb7c555990f8-Abstract-Conference.html}
\BIBentrySTDinterwordspacing

\bibitem{GaoGZMFZLQ24}
\BIBentryALTinterwordspacing
P.~Gao, S.~Geng, R.~Zhang, T.~Ma, R.~Fang, Y.~Zhang, H.~Li, and Y.~Qiao, ``Clip-adapter: Better vision-language models with feature adapters,'' \emph{Int. J. Comput. Vis.}, vol. 132, no.~2, pp. 581--595, 2024. [Online]. Available: \url{https://doi.org/10.1007/s11263-023-01891-x}
\BIBentrySTDinterwordspacing

\bibitem{carlini2017towards}
N.~Carlini and D.~Wagner, ``Towards evaluating the robustness of neural networks,'' in \emph{IEEE Symposium on Security and Privacy}, 2017, pp. 39--57.

\bibitem{GoodfellowSS14}
I.~J. Goodfellow, J.~Shlens, and C.~Szegedy, ``Explaining and harnessing adversarial examples,'' in \emph{International Conference on Learning Representations}, Y.~Bengio and Y.~LeCun, Eds., 2015.

\bibitem{zhu2023improving}
K.~Zhu, X.~Hu, J.~Wang, X.~Xie, and G.~Yang, ``Improving generalization of adversarial training via robust critical fine-tuning,'' in \emph{International Conference on Computer Vision}, 2023.

\bibitem{DBLP:conf/cvpr/MolchanovMTFK19}
\BIBentryALTinterwordspacing
P.~Molchanov, A.~Mallya, S.~Tyree, I.~Frosio, and J.~Kautz, ``Importance estimation for neural network pruning,'' in \emph{Proceedings of the IEEE/CVF conference on computer vision and pattern recognition}, 2019, pp. 11\,264--11\,272. [Online]. Available: \url{http://openaccess.thecvf.com/content\_CVPR\_2019/html/Molchanov\_Importance\_Estimation\_for\_Neural\_Network\_Pruning\_CVPR\_2019\_paper.html}
\BIBentrySTDinterwordspacing

\bibitem{DBLP:conf/iclr/CaiZH19}
\BIBentryALTinterwordspacing
H.~Cai, L.~Zhu, and S.~Han, ``Proxylessnas: Direct neural architecture search on target task and hardware,'' in \emph{International Conference on Learning Representations}, 2019. [Online]. Available: \url{https://openreview.net/forum?id=HylVB3AqYm}
\BIBentrySTDinterwordspacing

\bibitem{peft}
S.~Mangrulkar, S.~Gugger, L.~Debut, Y.~Belkada, S.~Paul, and B.~Bossan, ``Peft: State-of-the-art parameter-efficient fine-tuning methods,'' \url{https://github.com/huggingface/peft}, 2022.

\bibitem{DBLP:journals/corr/abs-2402-16843}
M.~Zhong, Y.~Shen, S.~Wang, Y.~Lu, Y.~Jiao, S.~Ouyang, D.~Yu, J.~Han, and W.~Chen, ``Multi-lora composition for image generation,'' \emph{arXiv preprint arXiv:2402.16843}, 2024.

\bibitem{ding2023parameter}
N.~Ding, Y.~Qin, G.~Yang, F.~Wei, Z.~Yang, Y.~Su, S.~Hu, Y.~Chen, C.-M. Chan, W.~Chen \emph{et~al.}, ``Parameter-efficient fine-tuning of large-scale pre-trained language models,'' \emph{Nature Machine Intelligence}, vol.~5, no.~3, pp. 220--235, 2023.

\bibitem{krizhevsky2009learning}
A.~Krizhevsky, G.~Hinton \emph{et~al.}, ``Learning multiple layers of features from tiny images,'' 2009.

\bibitem{RussakovskyDSKS15}
\BIBentryALTinterwordspacing
O.~Russakovsky, J.~Deng, H.~Su, J.~Krause \emph{et~al.}, ``Imagenet large scale visual recognition challenge,'' \emph{Int. J. Comput. Vis.}, vol. 115, no.~3, pp. 211--252, 2015. [Online]. Available: \url{https://doi.org/10.1007/s11263-015-0816-y}
\BIBentrySTDinterwordspacing

\bibitem{liu2021swin}
Z.~Liu, Y.~Lin, Y.~Cao, H.~Hu, Y.~Wei, Z.~Zhang, S.~Lin, and B.~Guo, ``Swin transformer: Hierarchical vision transformer using shifted windows,'' in \emph{Proceedings of the IEEE/CVF international conference on computer vision}, 2021, pp. 10\,012--10\,022.

\bibitem{DBLP:conf/icml/Croce020a}
\BIBentryALTinterwordspacing
F.~Croce and M.~Hein, ``Reliable evaluation of adversarial robustness with an ensemble of diverse parameter-free attacks,'' in \emph{Proceedings of Machine Learning Research}, ser. Proceedings of Machine Learning Research, vol. 119, 2020, pp. 2206--2216. [Online]. Available: \url{http://proceedings.mlr.press/v119/croce20b.html}
\BIBentrySTDinterwordspacing

\bibitem{wang2023paramet}
\BIBentryALTinterwordspacing
H.~Wang, X.~Yang, J.~Chang, D.~Jin, J.~Sun, S.~Zhang, X.~Luo, and Q.~Tian, ``Parameter-efficient tuning of large-scale multimodal foundation model,'' 2023. [Online]. Available: \url{https://arxiv.org/abs/2305.08381}
\BIBentrySTDinterwordspacing

\bibitem{abs-2410-05951}
\BIBentryALTinterwordspacing
K.~Lv, H.~Cao, K.~Tu, Y.~Xu, Z.~Zhang, X.~Ding, and Y.~Wang, ``Hyper adversarial tuning for boosting adversarial robustness of pretrained large vision models,'' \emph{CoRR}, vol. abs/2410.05951, 2024. [Online]. Available: \url{https://doi.org/10.48550/arXiv.2410.05951}
\BIBentrySTDinterwordspacing

\bibitem{loshchilov2017fixing}
I.~Loshchilov, F.~Hutter \emph{et~al.}, ``Fixing weight decay regularization in adam,'' \emph{arXiv preprint arXiv:1711.05101}, vol.~5, 2017.

\bibitem{0001DMPHGSCGAJB23}
M.~Dehghani, J.~Djolonga, B.~Mustafa, P.~Padlewski, J.~Heek \emph{et~al.}, ``Scaling vision transformers to 22 billion parameters,'' in \emph{International Conference on Machine Learning}.\hskip 1em plus 0.5em minus 0.4em\relax PMLR, 2023, pp. 7480--7512.

\bibitem{abs-2408-03326}
\BIBentryALTinterwordspacing
B.~Li, Y.~Zhang, D.~Guo, R.~Zhang \emph{et~al.}, ``Llava-onevision: Easy visual task transfer,'' \emph{CoRR}, vol. abs/2408.03326, 2024. [Online]. Available: \url{https://doi.org/10.48550/arXiv.2408.03326}
\BIBentrySTDinterwordspacing

\end{thebibliography}
}

\clearpage
\setcounter{page}{1}
\maketitlesupplementary

\section{Algorithm of CAAT}
\label{sec:rationale}
A detailed description of the CAAT algorithm is provided in Algorithm.\ref{alg:2}.
\begin{algorithm}
\caption{Criticality-Aware Adversarial Training }\label{alg:2}
\begin{algorithmic}[1]
\Require Pre-trained ViT model with network parameter $\theta$, adversarial dataset $\mathcal{D}_{adv}$, adversarial training iteration steps $T$, learning rate $\gamma$.
\Ensure The adversarial trained model weights $\theta^*_{AT}.$
\State Initialize criticality set: $\mathcal{C}=\left \{ 0 \right \}^N $.
\State \textbf{Step 1}: Calculate RPC for each parameter
\For{every parameter weight $\theta^j \in \theta$} 
\State{Calculate RPC of $\theta^j$ using Algorithm.\ref{alg:cap}.} 
\EndFor
\State \textbf{Step 2}: Allocate trainable adapter and AT.
\State Select the module whose RPC value exceeds the threshold to generate the binary mask matrix $M$ as shown in Eq.\ref{mask}.
\For{t =1,$\ldots$, $T$} \Comment{AT for $T$ epochs}
\For{Batch $\mathcal{B}^{adv} \in \mathcal{D}_{adv}$} 
\State{Calculate Loss $\rho(\theta, \mathcal{B}^{adv})$} 
\State{Calculate gradients $\Theta' \gets \Theta - \epsilon \mathbf{g} _{\Theta} \odot M$}
\EndFor
\EndFor
\State \textbf{Return}: The adversarial trained model weights $\theta^*_{AT}.$
\end{algorithmic}
\end{algorithm}
\section{Training Details}

\subsection{Experiment Code}
The source code is available at \url{https://anonymous.4open.science/r/CAAT-CF86}.
\section{Experiment Results}
\subsection{Effect of number of adversarial samples}
 We also investigate the effect of varying the number of adversarial samples used to calculate parameter criticality. Results in Table. \ref{tab:ablation_number} show that the robustness accuracy increases slightly from 200 to 600 samples and plateaus thereafter. This suggests that the robustness accuracy is not highly sensitive to the number of training samples. Calculating parameter criticality with 800 adversarial samples is efficient, with CAAT taking only 8 seconds per forward propagation.
\begin{table}[h]
	\centering
 \caption{Effect of the number of adversarial samples used to calculate parameter criticality. The best result is in \textbf{bold}.\label{tab:ablation_number}}
 \begin{tabular}{l|cccc}
		\hline
		Sample Num. & 200 & 400 & 600 & 800 \\ \hline
		PGD-10 Acc. & 50.10 & 50.13 & \textbf{50.15} &  \textbf{50.15} \\ \hline
	\end{tabular}
\end{table}

\section{Ethical Impact}
We propose Criticality-Aware Adversarial Training (CAAT), a method aimed at achieving parameter-efficient adversarial training. It is important to note that enhancing robustness is not the primary objective of this work. While our research currently focuses on ViT with relatively small parameter sizes, we plan to extend our methods to larger models. We believe that advancing robust research on large models, particularly LLM, will enhance the reliability and trustworthiness of LLM systems. A deeper understanding of the vulnerabilities in LLMs can enable their more effective utilization, especially in scenarios requiring high reliability.

\begin{table*}[t]
\small
	\centering
 \caption{Comparisons of different robust fine-tuning methods using a pre-trained Swin-B \cite{liu2021swin} backbone across various datasets.  The standard adversarial training is highlighted with a gray background. Additionally, the top-1 (\%) accuracy is reported in \textbf{bold}, and the top-2 (\%) accuracy is reported in \underline{underline}. \label{table:table_swin}}
	\begin{tabular}{c | c |c c |c  c c c c  }
\toprule[0.15em]
 Dataset & Method & \makecell[c]{Tuned params\\(M) } & \makecell[c]{Tuned / Total \\(\%) } & \makecell[c]{Clean Acc \\(\%) } & \makecell[c]{CW-20 \\(\%) } & \makecell[c]{PGD-10 \\(\%) } &  \makecell[c]{AutoAttack \\(\%) } \\
\midrule
\multirow{7}[4]{*}{CIFAR-10} &\cellcolor{mygray} \textsc{Full}  &\cellcolor{mygray} 86.69  & \cellcolor{mygray}100  &\cellcolor{mygray} 81.28    & \cellcolor{mygray}48.66  &\cellcolor{mygray} 50.69  &\cellcolor{mygray} 46.27     \\
\cmidrule{2-8}
 & \textsc{Adapter-8} & 1.72 & 1.98 & 75.68  & 42.10 & 38.85 & 39.37   \\
   & \textsc{ Adapter-32} & 2.00     & 2.31      & 76.53    & 42.53  & 39.88     & 39.47  \\
 & \textsc{LoRA-16} & 3.15 & 3.63& 80.84  & 45.73 & 46.59 & 43.81  \\
 & \textsc{FullLoRA-AT} & 3.17 & 3.66& 82.51  & 45.75 & 47.58 & \underline{44.27}  \\
\cmidrule{2-8}
  & \cellcolor[HTML]{FAEBD7} \textsc{CAAT-Adapter}  & \cellcolor[HTML]{FAEBD7}1.15 & \cellcolor[HTML]{FAEBD7}1.33 & \cellcolor[HTML]{FAEBD7}\underline{82.89} & \cellcolor[HTML]{FAEBD7}\underline{46.62} & \cellcolor[HTML]{FAEBD7}\underline{47.72} & \cellcolor[HTML]{FAEBD7}43.95   \\
  & \cellcolor[HTML]{FAEBD7} \textsc{CAAT-LoRA}  &\cellcolor[HTML]{FAEBD7} 1.25 &\cellcolor[HTML]{FAEBD7} 1.44 & \cellcolor[HTML]{FAEBD7}\textbf{83.57} &\cellcolor[HTML]{FAEBD7} \textbf{47.14} &\cellcolor[HTML]{FAEBD7} \textbf{48.15} &\cellcolor[HTML]{FAEBD7} \textbf{44.52}   \\
\midrule
\multirow{7}[4]{*}{CIFAR-100} &\cellcolor{mygray} \textsc{Full}  &\cellcolor{mygray} 86.76  & \cellcolor{mygray}100  &\cellcolor{mygray} 59.93    & \cellcolor{mygray}28.02  &\cellcolor{mygray} 30.48  &\cellcolor{mygray} 26.27     \\
\cmidrule{2-8}
 & \textsc{Adapter-8} & 1.72 & 1.98 & 50.52  & 20.52 & 21.06 & 20.22   \\
   & \textsc{ Adapter-32} & 2.00     & 2.31      & 51.77    & 20.18  & 21.72     & 21.37  \\
 & \textsc{LoRA-16} & 3.15 & 3.63& 58.32  & 24.77 & 26.19 & 21.82  \\
 & \textsc{FullLoRA-AT} & 3.17 & 3.65& 58.60  & 25.00 & 26.39 & 22.18  \\
\cmidrule{2-8}
  & \cellcolor[HTML]{FAEBD7} \textsc{CAAT-Adapter}  & \cellcolor[HTML]{FAEBD7}1.17 & \cellcolor[HTML]{FAEBD7}1.35 & \cellcolor[HTML]{FAEBD7}\underline{58.66} & \cellcolor[HTML]{FAEBD7}\underline{25.10} & \cellcolor[HTML]{FAEBD7}\underline{26.95} & \cellcolor[HTML]{FAEBD7}\underline{22.34}   \\
  & \cellcolor[HTML]{FAEBD7} \textsc{CAAT-LoRA}  &\cellcolor[HTML]{FAEBD7} 1.26 &\cellcolor[HTML]{FAEBD7} 1.45 & \cellcolor[HTML]{FAEBD7}\textbf{59.73} &\cellcolor[HTML]{FAEBD7} \textbf{25.63} &\cellcolor[HTML]{FAEBD7} \textbf{28.57} &\cellcolor[HTML]{FAEBD7} \textbf{24.94}   \\

\bottomrule[0.1em]
\end{tabular}
\end{table*}

\end{document}